\documentclass[10pt]{article}
\usepackage[preprint]{tmlr}

\usepackage{comment}
\excludecomment{draft}
\excludecomment{nextversion}

\usepackage[utf8]{inputenc} %
\usepackage[T1]{fontenc}    %
\usepackage{hyperref}       %
\usepackage{booktabs}       %

\usepackage{natbib}
\usepackage{babel}

\usepackage{amsthm}
\usepackage{amssymb}
\usepackage{amsfonts} %
\usepackage{thmtools,thm-restate}
\usepackage{mathrsfs}
\usepackage{mathtools} %
\usepackage{mathcommand}
\usepackage{nicefrac}       %
\usepackage{etoolbox}
\usepackage[bb=dsserif]{mathalpha}
\usepackage{cleveref}
\usepackage{textcomp}

\usepackage{tikz}

\usepackage{pifont}
\usepackage{graphicx}
\usepackage{booktabs} %
\usepackage{fancyhdr}
\usepackage{multirow}

\usepackage{url}            %
\usepackage{microtype}
\usepackage{graphicx}
\usepackage{subcaption} 
\usepackage{placeins} %

\usepackage{enumitem}
\setitemize{noitemsep,topsep=0pt,parsep=0pt,partopsep=0pt}
\setenumerate{noitemsep,topsep=0pt,parsep=0pt,partopsep=0pt}

\graphicspath{{./figs/}}

\providecommand\given{\MidSymbol[\vert]}

\newcommand\MidSymbol[1][]{%
\nonscript\:#1
\allowbreak
\nonscript\:
\mathopen{}}

\DeclareMathOperator{\opVar}{Var}

\DeclarePairedDelimiterXPP{\Var}[2]{\opVar_{#1}}{[}{]}{}{%
    \renewcommand\given{\MidSymbol[\delimsize\vert]}
    \ifblank{#2}{\:\cdot\:}{#2}
}

\DeclareMathOperator{\opExpectation}{\mathbb{E}}

\DeclarePairedDelimiterXPP{\implicitE}[1]{\opExpectation}{[}{]}{}{%
    \renewcommand\given{\MidSymbol[\delimsize\vert]}
    \ifblank{#1}{\:\cdot\:}{#1}
}

\DeclarePairedDelimiterXPP{\E}[2]{\opExpectation_{#1}}{[}{]}{}{%
    \renewcommand\given{\MidSymbol[\delimsize\vert]}
    \ifblank{#2}{\:\cdot\:}{#2}
}

\DeclarePairedDelimiterXPP{\indicator}[1]{\mathbb{1}}{\{}{\}}{}{%
    \ifblank{#1}{\:\cdot\:}{#1}
}

\DeclareMathOperator{\opInformationContent}{h}
\DeclarePairedDelimiterXPP{\ICof}[1]{\opInformationContent}{(}{)}{}{%
    \ifblank{#1}{\:\cdot\:}{#1}
}

\DeclareMathOperator{\opEntropy}{H}
\DeclarePairedDelimiterXPP{\Hof}[1]{\opEntropy}{[}{]}{}{%
    \renewcommand\given{\MidSymbol[\delimsize\vert]}
    \ifblank{#1}{\:\cdot\:}{#1}
}
\DeclarePairedDelimiterXPP{\xHof}[1]{\opEntropy}{(}{)}{}{%
    \ifblank{#1}{\:\cdot\:}{#1}
}

\DeclareMathOperator{\opMI}{I}
\DeclarePairedDelimiterXPP{\MIof}[1]{\opMI}{[}{]}{}{%
    \renewcommand\given{\MidSymbol[\delimsize\vert]}
    \ifblank{#1}{\:\cdot\:}{#1}
}

\DeclareMathOperator{\opTC}{TC}
\DeclarePairedDelimiterXPP{\TCof}[1]{\opTC}{[}{]}{}{%
    \renewcommand\given{\MidSymbol[\delimsize\vert]}
    \ifblank{#1}{\:\cdot\:}{#1}
}

\DeclarePairedDelimiterXPP{\CrossEntropy}[2]{\opEntropy}{(}{)}{}{%
    \ifblank{#1#2}{\:\cdot\: \MidSymbol[\delimsize\Vert] \:\cdot\:}{#1 \MidSymbol[\delimsize\Vert] #2}
}

\DeclareMathOperator{\opKale}{D_\mathrm{KL}}
\DeclarePairedDelimiterXPP{\Kale}[2]{\opKale}{(}{)}{}{%
    \ifblank{#1#2}{\:\cdot\: \MidSymbol[\delimsize\Vert] \:\cdot\:}{#1 \MidSymbol[\delimsize\Vert] #2}
}

\DeclareMathOperator{\opp}{p}
\DeclarePairedDelimiterXPP{\pof}[1]{\opp}{(}{)}{}{%
    \renewcommand\given{\MidSymbol[\delimsize\vert]}
    \ifblank{#1}{\:\cdot\:}{#1}
}

\DeclarePairedDelimiterXPP{\hpof}[1]{\hat{\opp}}{(}{)}{}{%
    \renewcommand\given{\MidSymbol[\delimsize\vert]}
    \ifblank{#1}{\:\cdot\:}{#1}
}

\DeclarePairedDelimiterXPP{\pcof}[2]{\opp_{#1}}{(}{)}{}{%
    \renewcommand\given{\MidSymbol[\delimsize\vert]}
    \ifblank{#2}{\:\cdot\:}{#2}
}

\DeclarePairedDelimiterXPP{\pfor}[1]{\opp}{\lbrack}{\rbrack}{}{%
    \renewcommand\given{\MidSymbol[\delimsize\vert]}
    \ifblank{#1}{\:\cdot\:}{#1}
}

\DeclarePairedDelimiterXPP{\pcfor}[2]{\opp_{#1}}{\lbrack}{\rbrack}{}{%
    \renewcommand\given{\MidSymbol[\delimsize\vert]}
    \ifblank{#2}{\:\cdot\:}{#2}
}

\DeclarePairedDelimiterXPP{\hpcof}[2]{\hat{\opp}_{#1}}{(}{)}{}{%
    \renewcommand\given{\MidSymbol[\delimsize\vert]}
    \ifblank{#2}{\:\cdot\:}{#2}
}

\DeclareMathOperator{\opq}{q}
\DeclarePairedDelimiterXPP{\qof}[1]{\opq}{(}{)}{}{%
    \renewcommand\given{\MidSymbol[\delimsize\vert]}
    \ifblank{#1}{\:\cdot\:}{#1}
}

\DeclarePairedDelimiterXPP{\qcof}[2]{\opq_{#1}}{(}{)}{}{%
    \renewcommand\given{\MidSymbol[\delimsize\vert]}
    \ifblank{#2}{\:\cdot\:}{#2}
}

\DeclarePairedDelimiterXPP{\qcpof}[2]{\opq'_{#1}}{(}{)}{}{%
    \renewcommand\given{\MidSymbol[\delimsize\vert]}
    \ifblank{#2}{\:\cdot\:}{#2}
}

\DeclarePairedDelimiterXPP{\varHof}[2]{\opEntropy_{\ifblank{#1}{\:\cdot\:}{#1}}}{[}{]}{}{%
    \renewcommand\given{\MidSymbol[\delimsize\vert]}
    \ifblank{#2}{\:\cdot\:}{#2}
}

\DeclarePairedDelimiterXPP{\xvarHof}[2]{\opEntropy_{\ifblank{#1}{\:\cdot\:}{#1}}}{(}{)}{}{%
    \renewcommand\given{\MidSymbol[\delimsize\vert]}
    \ifblank{#2}{\:\cdot\:}{#2}
}

\DeclarePairedDelimiterXPP{\varMIof}[2]{\opMI_{\ifblank{#1}{\:\cdot\:}{#1}}}{[}{]}{}{%
    \renewcommand\given{\MidSymbol[\delimsize\vert]}
    \ifblank{#2}{\:\cdot\:}{#2}
}

\DeclarePairedDelimiterXPP{\aICof}[1]{\opInformationContent'}{(}{)}{}{%
    \ifblank{#1}{\:\cdot\:}{#1}
}

\DeclarePairedDelimiterXPP{\aHof}[1]{\opEntropy'}{[}{]}{}{%
    \renewcommand\given{\MidSymbol[\delimsize\vert]}
    \ifblank{#1}{\:\cdot\:}{#1}
}
\DeclarePairedDelimiterXPP{\axHof}[1]{\opEntropy'}{(}{)}{}{%
    \ifblank{#1}{\:\cdot\:}{#1}
}

\DeclarePairedDelimiterXPP{\aMIof}[1]{\opMI'}{[}{]}{}{%
    \renewcommand\given{\MidSymbol[\delimsize\vert]}
    \ifblank{#1}{\:\cdot\:}{#1}
}

\DeclarePairedDelimiterXPP{\aCrossEntropy}[2]{\opEntropy'}{(}{)}{}{%
    \ifblank{#1#2}{\:\cdot\: \MidSymbol[\delimsize\Vert] \:\cdot\:}{#1 \MidSymbol[\delimsize\Vert] #2}
}

\newcommand{\Dataset}{\mathcal{D}}

\newcommand{\Dtrain}{{\Dataset_\text{train}}}

\newcommand{\pdataof}[1]{\hpcof{\text{data}}{#1}}

\newcommand{\pwof}[1]{\pcof{}{#1}}
\newcommand{\qwof}[1]{\qcof{}{#1}}

\newcommand{\w}{\omega}
\newcommand{\W}{\Omega}

\DeclareMathOperator*{\argmax}{arg\,max}
\DeclareMathOperator*{\argmin}{arg\,min}

\newcommand{\xeW}{\varHof{\hat \opp_\text{train}\Vert\opq}}
\newcommand{\xeWprime}{\varHof{\hat \opp_\text{train}\Vert\opq'}}
\newcommand{\HofW}{\varHof{\opq}}
\newcommand{\MIofW}{\varMIof{\opq}}
\newcommand{\xmiW}{\varMIof{\hat \opp_\text{train}\Vert\opq}}
\newcommand{\TCW}{\opTC_{\opq}}

\newcommand{\andreas}[1]{}
\newcommand{\yarin}[1]{}
\newcommand{\forreviewers}[1]{}
\newcommand{\jannik}[1]{}

\begin{draft}
\renewcommand{\andreas}[1]{{\leavevmode\color{blue}{ \footnotesize AK} {\tiny says: }#1}}
\renewcommand{\yarin}[1]{{\leavevmode\color{orange}{ \footnotesize YG} {\tiny says: }#1}}
\renewcommand{\forreviewers}[1]{{\leavevmode\color{purple}{\footnotesize NOTE} {\tiny please: }#1}}
\renewcommand{\jannik}[1]{{\leavevmode\color{teal}{\footnotesize JK} {\tiny says: }#1}}
\end{draft}

\everypar{\looseness=-1}
\allowdisplaybreaks

\title{Marginal and Joint Cross-Entropies \& Predictives for Online Bayesian Inference, Active Learning, and Active Sampling}

\author{\name Andreas Kirsch \email andreas.kirsch@cs.ox.ac.uk \\
\name Jannik Kossen \email jannik.kossen@cs.ox.ac.uk \\
\name Yarin Gal \email yarin.gal@cs.ox.ac.uk \\
      \addr OATML, Department of Computer Science \\
      University of Oxford
      }

\def\month{MM}  %
\def\year{YYYY} %
\def\openreview{\url{https://openreview.net/forum?id=XXXX}} %

\begin{document}

\maketitle

\begin{abstract}
    Principled Bayesian deep learning (BDL) does not live up to its potential when we only focus on marginal predictive distributions (marginal predictives).
    Recent works have highlighted the importance of joint predictives for (Bayesian) sequential decision making from a theoretical and synthetic perspective. We provide additional practical arguments grounded in real-world applications for focusing on joint predictives: we discuss \emph{online Bayesian inference}, which would allow us to make predictions while taking into account additional data \emph{without retraining}, and we propose new challenging evaluation settings using active learning and active sampling.
    These settings are motivated by an examination of marginal and joint predictives, their respective cross-entropies, and their place in offline and online learning.
    They are more realistic than previously suggested ones, building on work by \citet{wen2021predictions} and \citet{osb2022evaluating}, and focus on evaluating the performance of approximate BNNs in an online supervised setting.
    Initial experiments, however, raise questions on the feasibility of these ideas in high-dimensional parameter spaces with current BDL inference techniques, and we suggest experiments that might help shed further light on the practicality of current research for these problems.
    Importantly, our work highlights previously unidentified gaps in current research and the need for better approximate joint predictives.
\end{abstract}

\section{Introduction}

Deep learning has seen tremendous success in recent years. Beyond deep ensembles \citep{lakshminarayanan2016simple}, however, more principled methods of deep learning that attempt to be (approximately) Bayesian, commonly referred to as (approximate) 
\emph{Bayesian Neural Networks (BNN)}, have arguably not lived up to their full potential \citep{ovadia2019trust, beluch2018power}.
This might be because the focus in their evaluation has been on marginal predictions $\qwof{y \given x}$, where they can only provide \emph{marginal}\textsuperscript{:D} improvements over unprincipled regular NNs.
Yet the strength of a Bayesian approach for deep learning might not solely lie in marginal predictions but in allowing for online learning via online Bayesian inference.

With `\emph{online Bayesian inference}', we refer to incorporating additional data into the posterior predictive \emph{without} retraining in the common sense, i.e.~by computing gradients and optimizing the model parameters further. 
This could offer important performance benefits for applications that would otherwise require repeated retraining like active learning and could have important implications for how we could use large supervised models in production: currently, they are seen as strictly static, however online Bayesian inference would allow them to dynamically adapt to new data on the fly.

Generally, the difference between an approximate BNN and a regular NN is that the former assumes a distribution $\qof{\w}$ over the model parameters $\w$, where $\qof{\w}$ approximates the Bayesian posterior $\pof{\w \given \Dtrain}$, which is the optimal distribution given prior information $\pof{\w}$ and training data $\Dtrain$:
$\qof{\w} \approx \pof{\w \given \Dtrain}.$

\textbf{Online Bayesian Inference.} To incorporate new data $\{y_i, x_i\} \sim \pdataof{y, x}^n$, via online Bayesian inference, we simply apply Bayes' theorem:
for a test point $x$, the predictive $\qwof{y \given x, y_{n}, x_{n}, \ldots, y_1, x_1}$ is proportional to its joint predictive. We obtain:
\begin{align}
    \qwof{y \given x, y_{n}, x_{n}, \ldots, y_1, x_1} &=
    \frac{\qwof{y, y_{n}, \ldots, y_1 \given x, x_{n}, \ldots, x_1}}{\qwof{y_{n} \ldots, y_1 \given x_{n}, \ldots, x_1}} 
    \label{eq:online_bayesian_inference}
    \\
    &\propto \qwof{y, y_{n}, \ldots, y_1 \given x, x_{n}, \ldots, x_1}. 
\end{align}
We can thus use a joint predictive $\qwof{y, y_1, \ldots, y_n \given x, x_1, \ldots, x_n}$ to incorporate fixed $\{y_i, x_i\}^n$ and make predictions for $x$ without explicit retraining. %

Hence, for online Bayesian inference, we require joint predictives, which only Bayesian methods can give us:\footnote{For consistent joint predictives, adhering to the chain rule of probability, a model needs to adhere to the ``Bayesian update rule'', i.e.~Bayes' theorem, and thus is Bayesian. See also \cref{eq:prob_chain_rule} later.} through BNNs in the parametric case or through Gaussian processes in the non-parameteric case, for example. This strongly contrasts with marginal predictives which can also be modelled by regular NNs.

Can we perform online Bayesian inference well for high-dimensional inputs and parameters using current approximate BNNs?
The quality of the resulting predictions crucially depends on the joint predictives. 
However, computing joint predictives can be challenging. For example, in Bayesian literature, the joint predictive of all samples in the training set marginalized over the prior distribution is just the well-known marginal likelihood, which can be used for model selection \citep{mackay2003information, lyle2020bayesian, fern2020marginal}, and is known to be difficult to estimate in high-dimensional parameter spaces \citep{lotfi2022bayesian}. 
Similar challenges can be expected for performing online Bayesian inference using approximate posterior distributions.

\textbf{Related Work.} %
\citet{osb2022evaluating,wen2021predictions,osb2021neural,osb2021epistemic} mention the importance of joint predictives in the context of combinatorial decision problems, sequential predictions and multi-armed bandits in low dimensions. Compared to these previous works, we explore important connections to supervised learning, e.g.~in active learning \citep{atlas1989training, settles2009active} and Bayesian optimal experimental design \citep{lindley1956measure, foster2022a}, and focus on online Bayesian inference in high dimensions.
In addition, we clarify that both marginal and joint cross-entropies have their use, and it is not the case that one is always preferable over the other. Simply put, we will argue that they capture different quantities that are separately useful in offline and online learning, and we will combine them to evaluate the performance of online Bayesian inference using approximate BNNs. We provide more details in \S\ref{sec:related_work}.

\textbf{Marginal Cross-Entropy.} As will become evident, the marginal cross-entropy for a fixed predictive model captures the expected performance (\emph{under log loss}) when the model does not adapt to data at test time.
For supervised tasks, the marginal cross-entropy is what is commonly referred to as cross-entropy loss and represents common practice: we obtain a fixed set of parameters by training the model on a training set, and we re-use these parameters at test time without any further updates.
The performance of the model does not change as it observes more test data, and there is no feedback loop of any sort.
In this \emph{offline learning} setting, the marginal cross-entropy is the right choice to estimate performance.

\textbf{Joint Cross-Entropy.} On the other hand, the joint cross-entropy for a predictive model captures the performance in a \emph{sequential learning} setting, where sequential model updates take place.
Here, the parameter distribution $\qof{\w}$ serves as a prior for online Bayesian inference. This fits the context used in \citet{wen2021predictions} in which the model makes a prediction for the next step, observes the outcome, and then updates the model (agent). 

\textbf{Applications \& Experiments.} We will also see that the joint predictive is important for data selection in active learning and active sampling as we detail in \S\ref{sec:active_learning_and_active_sampling}.
We further connect several recent works \citep{wen2021predictions,osb2022evaluating} to active learning and active sampling and present new realistic and challenging experimental settings.
Most importantly, we examine online Bayesian inference within these contexts as it allows us to avoid \emph{retraining across acquisitions}.

\textbf{Outline.} In \S\ref{sec:background}, we introduce the model setting; in \S\ref{sec:margial_and_joint_cross-entropy}, we investigate marginal and joint cross-entropies; in \S\ref{sec:active_learning_and_active_sampling}, we show the connections to active learning and active sampling; in \S\ref{subsec:joint_entropy_data_adaptation}, we look at online Bayesian inference; in \S\ref{sec:evaluations_applications}, we present evaluation settings and applications, including for active learning and online Bayesian inference; 
in \S\ref{sec:experiments}, we suggest experiments and detail initial results on MNIST using MC dropout; and finally discuss related work in \S\ref{sec:related_work}.

\section{Background}
In this section, we introduce our setting, revisit marginal and joint cross-entropies, examine the connection to active learning and active sampling, and explain online Bayesian inference in detail.

\subsection{Setting}
\label{sec:background}
In the following, we will assume an underlying parametric predictive model $\pof{y \given x, \w}$ for input samples $x$ with targets or labels $y$
with a prior parameter distribution $\pof{\w}$ over $\w$, i.e.~our model is Bayesian.
The marginal predictive $\pwof{y\given x}$ is then obtained by marginalizing over $\pof{\w}$:
$\pwof{y \given x} = \E{\pof{\w}}{\pof{y \given x, \w}}.$
Similarly, the joint predictive is obtained by marginalizing the joint $\pof{y_1, \ldots, y_n \given x_1, \ldots, x_n, \w}$ over $\pof{\w}$: $\pwof{y_1, \ldots, y_n \given x_1, \ldots, x_n} = \E{\pof{\w}}{\pof{y_1, \ldots, y_n \given x_1, \ldots, x_n, \w}} = \E{\pof{\w}}{\prod_i \pof{y_i \given x_i, \w}}.$

We are interested in using the posterior parameter distribution $\pof{\w \given \Dtrain}$ to obtain marginal or joint predictives $\pwof{\cdot \given \Dtrain}$ given training data $\Dtrain$, where we use Bayes' theorem to obtain the posterior given a prior distribution $\pof{\w}$.

In general, capturing the true posterior distribution $\pof{\w \given \Dtrain}$ is infeasible, however. We assume we have an approximate distribution $\qof{\w} \approx \pof{\w \given \Dtrain}$ that we use instead of the true posterior. 

For example, $\qof{\w}$ could be based on a deep ensemble \citep{lakshminarayanan2016simple} as a mixture of Dirac delta distributions positioned at the parameters of individually trained ensemble members, or it could be an MC dropout model that is trained using variational inference \citep{gal2015dropout}. As above, we use $\qwof{y \given x}$ to denote the predictions after marginalizing over $\qwof{\w}$: $\qwof{y \given x} = \E{\qof{\w}}{\pof{y \given x, \w}}.$ Note that the underlying discriminative model $\pof{y \given x, \w}$ stays the same---we only exchange the distribution over its parameters $\w$. 

\textbf{Notation.} We use notation from \citet{kirsch2021practical} for information-theoretic quantities. In particular, this means that we take expectations over random variables (capital letters) and condition on outcomes (lower-case letters): e.g.~for entropies $\Hof{A, b \given C, d} = \E{\pof{a, c \given b, d}}{- \log \pof{a, b \given c, d}}.$ 
As examples:
\begin{align}
    \Hof{x} &\coloneqq -\log \pof{x}, \\
    \Hof{X} &= \E{\pof{x}}{\Hof{x}}, \\
    \Hof{x \given y} &= -\log \pof{x \given y}, \\
    \Hof{x \given Y} &= \E{\pof{y \given x}}{\Hof{x \given y}}, \\
    \Hof{X \given Y} &= \E{\pof{x, y}}{\Hof{x \given y}}, \\
    \Hof{X, y} &= \E{\pof{x \given y}}{\Hof{x, y}},
\end{align}
and so on. We define the cross-entropy as (only one example shown):
\begin{align}
    \varHof{\opp \Vert \opq}{Y \given x} \coloneqq \E{\pof{y \given x}}{- \log \qof{y \given x}}.
\end{align}
Similarly,  the mutual information is consistently defined via:
\begin{align}
    \MIof{x ; y} &\coloneqq \Hof{x} - \Hof{x \given y}, \\
    \MIof{X ; y} &= \Hof{X} - \Hof{X \given y}, \\
    \MIof{x ; Y} &= \Hof{x} - \Hof{x \given Y}, \\
    \MIof{X ; Y} &= \Hof{X} - \Hof{X \given Y}.
\end{align}
This matches the well-known standard definitions and extends the notation to mixing random variables and outcomes. The mixed mutual information terms capture the notion of information gain and information-theoretic surprise, respectively---see \citet{kirsch2021practical} for more details. We will add the distribution as a subscript to disambiguate the notation: e.g.~$\varHof{\opq}{X}.$

\subsection{Marginal and Joint Cross-Entropy}
\label{sec:margial_and_joint_cross-entropy}
We start by comparing marginal and joint predictive cross-entropies and revisiting how they are useful for offline and online learning separately.

\textbf{Marginal Cross-Entropy.} Given an underlying, possibly empirical, data distribution $\pdataof{x,y}$, the \emph{marginal cross-entropy}\yarin{maybe both this and $\qwof{y\given x}$ in color?} is:
\begin{align}
    \xeW{Y \given X} &= \E{\pdataof{x, y}}{-\log \E{\qof{\w}}{\pof{y\given x, \w}}} \notag \\
    &= \E{\pdataof{x, y}}{-\log \qwof{y\given x}},
\end{align}
where we use $\xeW{Y \given X}$ to denote the cross-entropy.
This cross-entropy is the population loss when $\qof{\w}$ is not updated after seeing new samples. Each sample $x, y$ is treated independently. Hence, the marginal cross-entropy captures the expected performance in an \emph{offline learning} setting.

\textbf{Joint Cross-Entropy.} On the other hand, given an initial parameter distribution $\qof{\w}$ above, the joint cross-entropy measures how well the parameter distribution can \emph{adapt} to new data $\Dataset$. 
\jannik{perhaps you want to show the joint cross-entropy equation first once, before doing (4)-(7)?}

To show how this connects to joint cross-entropies, we can look at the joint cross-entropy of specific samples $\Dataset = \{y_i, x_i\}_{i=1}^n$ (without taking an expectation).
The joint cross-entropy for these specific samples is just the sum of (negative) log marginal likelihoods using the chain rule, where each $y_i$ is conditioned on all `previous' observations $\Dataset_{<i}$:
\begin{align}
    \xeW{y_1, \ldots, y_n \given x_1, \ldots, x_n} &= -\log \qwof{y_1, \ldots, y_n \given x_1, \ldots, x_n} \\
    &= -\log \prod_i \qwof{y_i \given x_i, y_{i-1}, x_{i-1}, \ldots, y_1, x_1} \\
    &= -\sum_i \log \qwof{y_i \given x_i, \Dataset_{<i}} \label{eq:prob_chain_rule} \\
    &= \sum_i \xeW{y_i \given x_i, \Dataset_{<i}},
\end{align}
where $\Dataset_{<i}$ denotes ``$y_{i-1}, x_{i-1}, \ldots, y_1, x_1$'', and the marginal predictive is:
\begin{math}
    \qwof{y_i \given x_i, \Dataset_{<i}} = \E{\qof{\w \given \Dataset_{<i}}}{\pof{y_i \given x_i, \w}}.
\end{math}
Semantically, we compute the following in each iteration of the sum: we \emph{update} the parameter posterior $\qof{\w \given \Dataset_{<i}}$, compute the losses for our predictions at outcomes $y_i$ for $x_i$, and then include $y_i, x_i$ in our observed data.
We will denote this as the \emph{online learning setting}.

When we are interested in the expected loss given arbitrary data, we can compute the following joint cross-entropy:
\newcommand{\OLL}[1]{OLL(#1)}
\begin{align}
    \OLL{n} &\coloneqq \E{(x_i, y_i)_i \sim \pdataof{x_i, y_i}^n}{-\log \qwof{y_1, \ldots, y_n \given x_1, \ldots x_n}} \\
    &= \E{(x_i, y_i)_i \sim \pdataof{x_i, y_i}^n}{\xeW{y_1, \ldots, y_n \given x_1, \ldots x_n}} \\
    &= \xeW{Y_1, \ldots, Y_n \given X_1, \ldots X_n},
\end{align}
where $OLL$ stands for ``online learning loss''.

\textbf{Connection to the Conditional Cross-Entropy Rate.} %
As an aside, if we let $n \to \infty$, we also have $\OLL{n} \to \infty$. This is not helpful, so instead we can look at the average: $\frac{1}{n} \OLL{n}$. In the limit, this average is just the \emph{cross-entropy rate}:
\begin{align}
    \xeW{\mathcal{Y} \given \mathcal{X}} \coloneqq \lim_{n \to \infty} \frac{1}{n} \xeW{Y_1, \ldots, Y_n \given X_1, \ldots X_n},
\end{align}
which we define analogously to the entropy rate in \citet{cover1991information}.\footnote{Cf.~the entropy rate, which is:
\begin{math}
    \Hof{\mathcal{X}} = \lim_{n \to \infty} \frac{1}{n} \Hof{X_1, \ldots X_n}.
\end{math}
}

\textbf{Summary.} The marginal cross-entropy is useful for offline learning as it predicts the performance of a fixed model on the data distribution. The joint cross-entropy is useful for online learning as it predicts the performance of a model as it adapts to additional data. Hence, it is important for sequential decision making where we often cannot afford to retrain the model dynamically. 

\subsection{Connection to Active Learning and Active Sampling}
\label{sec:active_learning_and_active_sampling}

Marginal and joint cross-entropies and the related predictives play an important role in active learning and active sampling as we will examine here.
In \emph{active learning} \citep{atlas1989training, settles2009active}, we do not know the outcome (label) for every sample and have a budget for the additional labels we can acquire for training from an unlabeled pool set. In \emph{active sampling} (also data subset selection or coreset selection) \citep{campbell2017automated,mirzasoleiman2019coresets, borsos2020coresets}, on the other hand, we have access to the labels, but we assume that we have a budget for the samples we can use to train the model on, and our goal is to pick the best subset of labeled samples.

In each case, we have a budget of $n$ additional samples we can condition on and we are usually interested in maximizing the performance of the model as end-goal, which is equivalent to finding $x_1,\ldots,x_n$ which minimize the following cross-entropy (assuming the usual cross-entropy loss):
\begin{align}
    \xeW{Y \given X, y_n, x_n, \ldots, y_1, x_1} \text{ or } \xeW{Y \given X, Y_n, x_n, \ldots, Y_1, x_1},
\end{align}
depending on whether we have access to the labels (active sampling) or not (active learning).

These objectives depend on a joint predictive via \cref{eq:online_bayesian_inference} but in expectation over $X$ and $Y$ sampled from the data distribution. As such it is a hybrid between the joint and marginal cross-entropy.

\textbf{Transductive Active Sampling.} %
If we have access to the labels $y_1, \ldots, y_n$, this leads to active sampling approaches, where we want to find the best samples to train on to increase model performance.
Using
\begin{align}
    \xmiW{Y ; y_n, \ldots, y_1 \given X, x_n, \ldots, x_1} = \xeW{Y \given X} - \xeW{Y \given X, y_n, x_n, \ldots, y_1, x_1},
\end{align}
where $\xmiW{}$ is a ``cross mutual information'', \emph{minimizing} above cross-conditional entropy is equivalent to \emph{maximizing} the cross-mutual information:
\begin{align}
    \argmin_{\{x_i\}_{1..n}} \xeW{Y \given X, y_n, x_n, \ldots, y_1, x_1} = \argmax_{\{x_i\}_{1..n}} \xmiW{Y ; y_n, \ldots, x_1 \given X, x_n, \ldots, x_1},
\end{align}
because $\xeW{Y \given X}$ is constant (independent of the $x_i$).

Approximations of this term lead to the \emph{reducible hold-out loss} in \citet{mindermann2021prioritized}. Note that this is a transductive approach \citep{yu2006active} because we examine performance in regards to samples $x, y$ from the data distribution.

\textbf{Transductive Active Learning.} %
If we do not have access to the labels, the Bayesian-optimal approach is to take the expectation using labels drawn from the joint predictive of the model \citep{lindley1956measure}. This leads to minimizing conditional entropies or, equivalently, maximizing mutual information terms \citep{mackay1992information, McCallum1998EmployingEA, yu2006active, wang2020marginal, kirsch2021test}. Assuming we have access to no labels beyond the current training set at all, we have the following objectives:
\begin{align}
    \argmin_{\{x_i\}_{1..n}} \HofW{Y \given X, Y_n, x_n, \ldots, Y_1, x_1} = \argmax_{\{x_i\}_{1..n}} \MIofW{Y ; Y_n, \ldots, Y_1 \given X, x_n, \ldots, x_1}.
\end{align}
The latter is exactly the Expected Predictive Information Gain (EPIG) acquisition function from \citet{kirsch2021test}.\footnote{EPIG is formulated using an unlabeled evaluation set that specifies the target domain. Here, we use the pool set itself as an evaluation set to simplify the exposition.}

\textbf{Active Learning.} %
Many active learning approaches, especially the ones that are not transductive, maximize the \emph{expected information gain}, also called BALD \citep{houlsby2011bayesian}, as an acquisition function---or, equivalently, minimize the model uncertainty.  
The corresponding information-theoretic expression is $\MIofW{\W; Y \given x}$, and in the batch case, we have:
\begin{align}
    \argmax_{\{x_i\}_{1..n}} \MIofW{\W ; Y_n, \ldots, Y_1 \given x_n, \ldots, x_1} &=
    \argmax_{\{x_i\}_{1..n}} \HofW{Y_n, \ldots, Y_1 \given x_n, \ldots, x_1} - \HofW{Y_n, \ldots, Y_1 \given x_n, \ldots, x_1, \W} \notag \\
    &= \argmax_{\{x_i\}_{1..n}} \HofW{Y_n, \ldots, Y_1 \given x_n, \ldots, x_1} - \sum_i \HofW{Y_i, \given x_i, \W}, 
\end{align}
where the last step follows from the fact that predictions factorize conditioned on the model parameters \citep{kirsch2019batchbald}.
The corresponding active sampling acquisition function $\MIofW{\W ; y_n, \ldots, y_1 \given x_n, \ldots, x_1}$, which quantifies the information gain, is examined further in \citet{kirsch2021practical}.

\textbf{Summary.} All of the above objectives depend on the joint predictive in some way to compute acquisition scores without explicit retraining. However, it is common to update models after acquiring a set of labels to take into account the newly acquired data.

\subsection{Online Bayesian Inference}
\label{subsec:joint_entropy_data_adaptation}
\andreas{check in the osband paper if they have any paper we can cite here for online learning/sequential decision making}

To see what we mean by incorporating new data, assume we have sampled $n$ additional points $x_i, y_i \sim \pdataof{x_i, y_i}$. 
Traditionally, we would now update the posterior approximation $\qof{w}$ to take this new data into account for our predictions at future test points. However, this can be prohibitively expensive---especially in applications that require frequent retraining. Instead, online Bayesian inference allows Bayesian models to adapt their predictions without explicitly updating the posterior approximation.

Following \cref{eq:online_bayesian_inference}, for a test point $x$, the predictive $\qwof{y \given x, y_{n}, x_{n}, \ldots, y_1, x_1}$ is proportional to the joint predictive:
\begin{align}
    \qwof{y \given x, y_{n}, x_{n}, \ldots, y_1, x_1} &=
    \frac{\qwof{y, y_{n}, \ldots, y_1 \given x, x_{n}, \ldots, x_1}}{\qwof{y_{n} \ldots, y_1 \given x_{n}, \ldots, x_1}} \\
    &\propto \qwof{y, y_{n}, \ldots, y_1 \given x, x_{n}, \ldots, x_1},
\end{align}
since the normalization constant $\qwof{y_{n} \ldots, y_1 \given x_{n}, \ldots, x_1}$ is independent of $y$ and $x$.
Hence, this allows us to make predictions that take into account new data \emph{without explicit retraining} by simply computing the joint predictive of the test point and newly observed data.

We refer to this as \emph{online Bayesian inference (OBI)}. While this inference is precisely Bayesian, $\qof{\w}$ is commonly only an approximate posterior, and thus the quality of this inference depends on the properties of the approximation and how we estimate the joint predictive. 

The simplest approach to estimate the joint predictive is via sampling, which applies to e.g.~Monte-Carlo dropout, deep ensembles, and deep ensembles with prior functions \citep{gal2015dropout, lakshminarayanan2016simple, osb2018randomized}, by factorizing the joint:
\begin{align}
    \qwof{y, y_{n}, \ldots, y_1 \given x, x_{n}, \ldots, x_1} &= \E{\qof{\w}}{\pof{y, y_{n}, \ldots, y_1 \given x, x_{n}, \ldots, x_1, \w}} \\
    &= \E*{\qof{\w}}{\pof{y \given x, \w} \, \prod_{i=1}^n \pof{y_{i} \given x_{i}, \w}}.
\end{align}
Thus, if we draw fixed parameter samples $\w_j \sim \qof{\w}$, we can pre-compute $\prod_{i=1}^n \pof{y_{i} \given x_{i}, \w_j}$ for each $j$ and estimate the joint predictive. 

Finally, we can view $\prod_{i=1}^n \pof{y_{i} \given x_{i}, \w}$ as unnormalised importance weights:
\begin{align}
    \qof{\w} \, \prod_{i=1}^n \pof{y_{i} \given x_{i}, \w} \propto \qof{\w \given y_{n}, x_{n}, \ldots, y_1, x_1},
\end{align}
and hence, overall:
\begin{align}
    \E*{\qof{\w}}{\pof{y \given x, \w} \, \prod_{i=1}^n \pof{y_{i} \given x_{i}, \w}} \propto \E{\qof{\w \given y_{n}, x_{n}, \ldots, y_1, x_1}}{\pof{y \given x, \w}}. \label{eq:implicit_posterior_samples}
\end{align}

To evaluate the performance, we can use these predictions to compute the following marginal cross-entropy which incorporates the additional samples using OBI:
\begin{align}
    \xeW{Y \given X, y_n, x_n, \ldots, y_1, x_1},
\end{align}
where $X, Y$ are sampled from the data distribution. Comparing this entropy with the performance of a fully retrained model allows us to obtain a practical estimate for the quality of the approximate posterior $\qof{\w}$.

\section{New Evaluations \& Applications}
\label{sec:evaluations_applications}

We suggest new experimental settings that allow us to evaluate the quality of the joint predictive and compare it to ones suggested in prior work.

\subsection{Performance in Active Learning and Active Sampling Methods}
A conceptually simple set of downstream tasks is to evaluate the performance of the joint predictives in different active learning or active sampling settings using different approximate model architectures (e.g.~based on Epistemic Neural Networks \citet{osb2021epistemic} as abstraction). \citet{wang2020marginal} show that performance for transductive active learning is correlated to the quality of the joint predictive. Similarly, \citet{kirsch2019batchbald} found the joint predictive to perform much better in batch active learning than a factorized distribution relying on marginal predictives.

We suggest, similar to Repeated-MNIST \citep{kirsch2019batchbald}, to duplicate the underlying pool sets or to simply \emph{allow the same sample to be selected multiple times}. Importantly, approximate BNNs that do not provide good joint predictives will greedily select the same sample over and over again or degrade to uninformed data acquisitions. This avoids an issue pointed out by \citet{wang2020marginal} and \citet{osb2022evaluating} as we explain next.

\textbf{Connection to Total Correlation.} %
\citet{wang2020marginal} argue that the joint cross-entropy is dominated by the sum of the individual marginal cross-entropy scores. This is equivalent to saying that the total correlation between samples is negligible since the difference between the joint cross-entropy and its individual marginal cross-entropies for specific $y_i, x_i$ is just the \emph{total correlation}:
\begin{align}
    \TCW[y_1, \ldots, y_n \given x_1, \ldots, x_n] \coloneqq \sum_i \HofW{y_i \given x_i} - \HofW{y_1, \ldots, y_n \given x_1, \ldots, x_n}.
\end{align}
The total correlation measures the amount of information shared between the samples.

For random batches, the total correlation is, indeed, likely going to be negligible because most random batches are not very informative overall, and importantly, on curated datasets, they are most likely uncorrelated as it is unlikely that observing $y_i$, $x_i$ in the batch informs prediction for $y_j$, $x_j$ for most $j \not= i$ as curated datasets are usually as diverse as possible.

Only with increasing redundancy in the dataset, e.g. by duplicating samples like in Repeated-MNIST \citep{kirsch2019batchbald}, random batches will become more correlated on average and the total correlation larger. %

This setup is similar to the dyadic sampling proposed by \citet{osb2022evaluating} which repeatedly samples $y^j_i$ for $x_i$, with $i \in \{1,2\}$ and evaluates the joint predictive. 
However, this setting in essence only measures the ability of the approximate model to perform Bayesian updates on two fixed training samples at a time.
Hence, we suggest that a better adaption to evaluate joint predictives using active learning is to duplicate the dataset or to simply allow the same sample to be selected multiple times.

\subsection{Performance of Online Bayesian Inference} %
As a practical ``ground truth'', we can compare the performance of retrained models after acquiring additional samples with the performance of OBI as explained in \cref{subsec:joint_entropy_data_adaptation}.

A particular challenging scenario for OBI is to use acquisition sequences $x_1, y_1, \ldots x_T, y_T$ that were collected using active learning or active sampling on the dataset. We can evaluate OBI on models trained at different $\Dtrain_{,t} = \{y_i,x_i\}_{i=1}^t$ and increasing subsets of online data $\{y_i, x_i\}_{i=t+1}^T$ from these acquisition sequences. This scenario is particularly challenging for OBI because the sequence of acquisition is selected to result in large changes in the predictives and posterior distributions.

The average performance difference between OBI and fully retrained models across different training acquisition sequences will tell us how good a given joint predictive is for ``meaningful'' online learning.
The expectation is that for most approximate BNNs, OBI will quickly suffer from degraded performance compared to the retrained models.

That is, we compare the performance of OBI as we acquire new samples $x_i, y_i$ to an approximate BNN retrained with the same additional data for increasing $n$:
\begin{align}
    \xeW{Y \given X, y_n, x_n, \ldots, y_1, x_1} - \xeWprime{Y \given X}, 
    \label{eq:joint_chain_error}
\end{align}
where $\qcpof{}{\w} \approx \pof{\w \given y_n, x_n, \ldots, y_1, x_1, 
\Dtrain}$ is the parameter distribution of an (approximate) BNN after retraining with the additional $y_n, x_n, \ldots, y_1, x_1$. %

Ideally, we would compare to the predictions from the correct updated posterior distribution; however, this is infeasible in most practical scenarios.
Instead, when we use an approximate BNN $\qcpof{}{\w}$ that is similar to the one used for $\qof{\w}$, we can measure the practical degradation between OBI and retraining. Here, the ideal would be for OBI to behave exactly like a fully retrained model---even if the latter does not match exact Bayesian inference---as such an approach would be \emph{self-consistent}.
Note that with exact Bayesian inference, we would have $\qof{\w} = \qcpof{}{\w}$ and the above would be zero. %

\textbf{Comparison to Dyadic Sampling.} Unlike \citet{osb2022evaluating} which focuses on selecting labels for dyadic samples repeatedly, this experiment setting is both more practical and more insightful: active learning picks samples that are the most informative and are supposed to update the posterior the most. This is because, for informative samples, we would expect the changes in model predictions to be the largest. Hence, one could expect that these samples pose the most significant challenge to approximate BNNs and their joint predictives. Ideally, we would hope that OBI would keep up with retrained model, but this might prove to be challenging in high-dimensional scenarios.

\textbf{Sample Selection Bias.} %
The suggested evaluation is orthogonal to any sample selection bias that is added through the data acquisition process itself as we use the same training data at each step for both OBI and retraining in \cref{eq:joint_chain_error}.
Specifically, \citet{farquhar_statistical_2020} observed that active learning introduces a bias by sampling from the data distribution using an acquisition function and not uniformly.

\jannik{uff! really not sure about this one. the bias is in the learned parameters.

since OBI uses bayesian inference, while q(w) uses approximate inference, the biases might manifest very differently in the parameters and thus predictions, and thus might not cancel out at all!!

definitely check with seb!}

\subsection{Application: Active Learning with Online Bayesian Inference} 
In many settings, retraining models for when only few new samples are added is prohibitively expensive. This motivates batch active learning, where batches are acquired instead of individual samples. Expanding on this, when equipped with models that perform well under OBI, one could avoid retraining models when acquiring new data by using OBI. Only when OBI degrades, fully retraining will become necessary.

We can evaluate this both for individual acquisition as well as for batch acquisition.

\section{Experiments}
\label{sec:experiments}

\begin{figure}[tb]
    \centering
    \includegraphics[width=\textwidth]{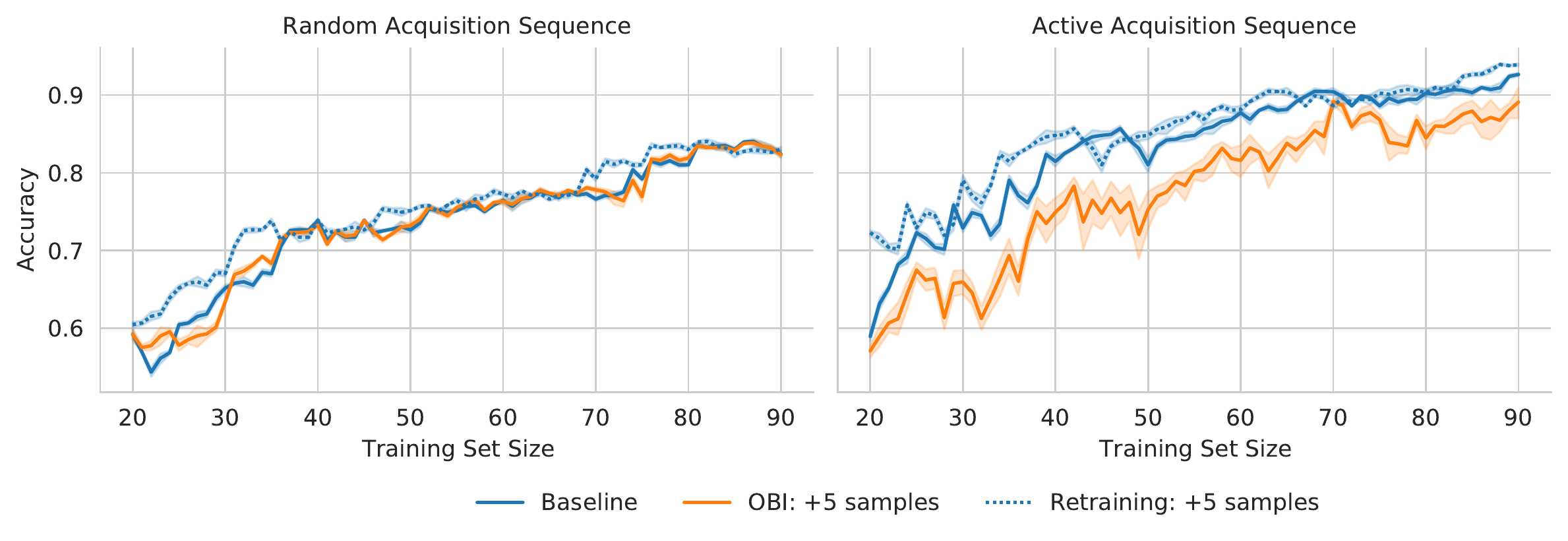}
    \includegraphics[width=\textwidth]{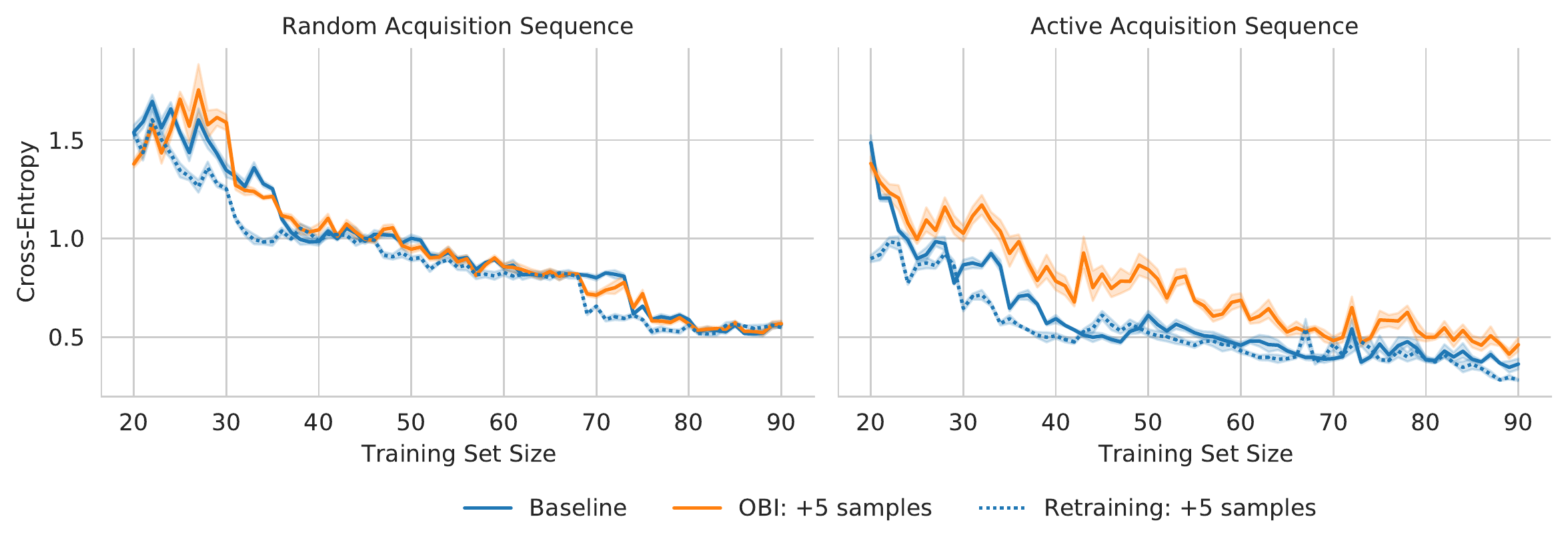}
    \caption{\emph{Comparison between online Bayesian inference and retraining for 5 additional samples on MNIST.} We compare the dynamics between using a random acquisition sequence and using an \emph{active} acquisition sequence, drawn using active sampling. While OBI does not appear to be significantly worse or better on the random acquisition sequence, it markedly deteriorates when using the active acquisition sequence. OBI struggles with the informative samples that change the posterior a lot: the active acquisition sequence reaches 90\% accuracy with just 70 samples, unlike the random one.}
    \label{fig:obi_mnist_comparison}
\end{figure}

\begin{table}[tb]
    \caption{\emph{Comparison between online Bayesian inference and retraining for 5 additional samples on MNIST.} OBI performs worse than the baseline (ie. not taking into account new samples at all) when using an active acquisition sequence. On the random acquisition sequence, it only performs just as well as not updating at all. (Mean for 5 model trials and 5 OBI trials for each.)}
    \centering
    \resizebox{\linewidth}{!}{%
    \begin{tabular}{lrrrr}
    \toprule
    {Baseline \textbf{vs}} & \multicolumn{2}{l}{OBI: +5 samples} & \multicolumn{2}{l}{Retraining: +5 samples} \\
    {Avg} & $\Delta$ Cross-Entropy (\textdownarrow) &  $\Delta$ Accuracy (\textuparrow) & $\Delta$ Cross-Entropy (\textdownarrow) & $\Delta$ Accuracy (\textuparrow) \\
    \midrule
    Active Acquisition Sequence &     0.16 & -5.7\% &    -0.062 &  2.0\% \\
    Random Acquisition Sequence &     $0.00$ &  $0.0\%$ &    -0.075 &  1.8\% \\
    \bottomrule
    \end{tabular}}
    \label{tab:obi_mnist_comparison}    
\end{table}

Following the newly suggested experimental settings, we want to run comparisons using different kinds of approximate BNNs in active learning and active sampling settings.
Moreover, given fixed acquisition sequences, determined using active learning and active sampling, we want to evaluate the difference between OBI and fully retrained models.

An initial experiment shows that OBI in high dimensions might not be feasible using simple Monte Carlo approximations of the expectations in parameter space. This is likely because of the much higher dimensionality of the problems we consider---especially in comparison to \citet{osb2021neural}.
\andreas{run a comparison to random acquisition sequences? and/or with deep ensembles? I could easily train 16 DEs}

Specifically, we use an acquisition sequence created using active sampling which achieves ~90\% accuracy on MNIST with just 70 samples and use the same model and training setup as \citet{kirsch2021practical}. After every new data point (starting from 20), we evaluate a retrained model using OBI with 5 additional data points and compare it to a fully retrained model with the same 5 additional data points as well as to a model that is not retrained at all. We train the models (5 trials) for each training set size and (bootstrap) sample 10000 MC dropout samples for OBI 5 times out of 20000 MC dropout samples using consistent MC dropout for each model \citep{kirsch2019batchbald, gal2015dropout} (5 sub-trials) to reduce the variance.

Ideally, when using OBI, we should recover the same performance as if we fully retrained the models using the additional data. However, as is visible in \Cref{fig:obi_mnist_comparison}, this is not the case when using the challenging acquisition sequence from active sampling.

\Cref{tab:obi_mnist_comparison} shows the average performance difference when using OBI with additional samples from the acquisition sequence and when fully retraining. In all cases, OBI performs worse than retraining. On the random acquisition sequence, it performs only as well as not updating at all, while on the active acquisition sequence, it always performs worse.

\section{Related Work}
\label{sec:related_work}
The most relevant works and indeed one of the inspirations for this work are \citet{wen2021predictions} and \citet{osb2022evaluating}, which are recommended reading. We see our work as a contribution that provides a different and differentiated position on the benefits of marginal versus joint predictives and respective cross-entropies as performance metrics and that puts greater focus on OBI.

Moreover, our suggested experimental settings expand on these prior works and draw attention to active learning and active sampling as more realistic use-cases. Measuring the error between OBI and retrained models expands on the experiments in \citet{wen2021predictions} while evaluating performance in active learning and active sampling on highly redundant datasets (allowing to reselect previously selected points) expands on the idea of dyadic sampling from \citet{osb2022evaluating}.

\newcommand{\dkl}[1]{d_{KL}^{#1}}
\newcommand{\dce}[1]{d_{CE}^{#1}}

Lastly, \citet{wen2021predictions} focus on the KL divergence between the exact Bayesian joint predictive and the joint predictive of an approximate Bayesian model for different numbers of samples in the joint.
Our suggested experimental settings focus on evaluating downstream tasks.

\citet{wang2020marginal} also examine the quality of joint predictives on low-dimensional datasets using a more synthetic evaluation method, the cross-normalized log-likelihood. They also evaluate the quality of joint predictives in regression settings using active learning experiments. Our evaluation settings from \S\ref{sec:evaluations_applications} extend these.

\section{Conclusion}

We have revisited the difference between marginal and joint cross-entropies and predictives, clarifying in which contexts either is appropriate: for offline learning, the marginal cross-entropy is the right choice to evaluate performance while for online learning, it is the joint cross-entropy. 
We have also shown how the joint predictive plays an important role in information-theoretic acquisition functions in active learning and active sampling.

Importantly, we argue that online Bayesian inference could provide many benefits and have proposed new more practical and challenging experimental settings which expand on prior art by using active learning and active sampling.

Given the results of the presented experiment, it is an open question how much better other sampling-based approaches can be when using high-dimensional parameter spaces. Especially deep ensembles which usually provide a much smaller ``sample count'' (i.e.~number of ensemble members) might not perform well under online Bayesian inference because the hypothesis space will be exhausted faster---even when the ensemble members are diverse. 

In a future revision, we will offer further experimental evaluation following \S\ref{sec:evaluations_applications}, e.g.~improving the quality of online Bayesian inference by studying higher quality posterior distributions such as those from HMC or efficient low-dimensional posterior approximations that might make parameter-space integrals tractable, and we will investigate if prior research into failures of Bayesian model averaging under model misspecification might provide further insights and paths to improvements \citep{Minka2002BayesianMA}.

\FloatBarrier

\section*{Acknowledgements}

The authors would like to thank the members of OATML in general for their feedback at various stages of the project. AK is supported by the UK EPSRC CDT in Autonomous Intelligent Machines and Systems (grant reference EP/L015897/1). JK is supported by New College Yeotown Scholarship.

\bibliographystyle{plainnat}
\bibliography{references}

\clearpage
\appendix

\end{document}